\documentclass[10pt,twocolumn,letterpaper]{article}

\usepackage[pagenumbers]{cvpr} %
\usepackage{bm}
\usepackage{multirow}
\usepackage{tabularx}
\usepackage{float}
\usepackage{lipsum}

\usepackage{breakcites} %
\usepackage{microtype} %

\definecolor{cvprblue}{rgb}{0.21,0.49,0.74}
\usepackage[pagebackref,breaklinks,colorlinks,allcolors=cvprblue]{hyperref}

\title{Continuous 3D Perception Model with Persistent State}

\author{Qianqian Wang$^{1,2*}$, Yifei Zhang$^{1*}$, Aleksander Holynski$^{1,2}$, Alexei A. Efros$^1$, Angjoo Kanazawa$^1$
\\[2ex]
\centering
 $^1$University of California, Berkeley  \hspace{3em} $^2$Google DeepMind
}

\newcommand{\It}{\bm{I}_t}
\newcommand{\ft}{\bm{F}_t}
\newcommand{\fts}{\bm{F}_t'}
\newcommand{\stpre}{\bm{s}_{t-1}}
\newcommand{\stcur}{\bm{s}_{t}}
\newcommand{\zts}{\bm{z}_t'}
\newcommand{\pmself}{\hat{\bm X}_t^\text{self}}
\newcommand{\pmcross}{\hat{\bm X}_t^\text{world}}
\newcommand{\confself}{\bm C_t^\text{self}}
\newcommand{\confcross}{\bm C_t^\text{world}}
\newcommand{\headself}{\mathrm{Head}_\text{self}}
\newcommand{\headcross}{\mathrm{Head}_\text{world}}
\newcommand{\headpose}{\mathrm{Head}_\text{pose}}
\newcommand{\raymap}{\bm R}
\newcommand{\fr}{\bm{F}_r}
\newcommand{\frs}{\bm{F}_r'}
\newcommand{\headcolor}{\mathrm{Head}_\text{color}}
\newcommand{\pmselfset}{\mathcal{\hat{X}}^\text{self}}
\newcommand{\pmcrossset}{\mathcal{\hat{X}}^\text{world}}
\newcommand{\pose}{\hat{\bm P}_t}
\newcommand{\predrgb}{\hat{\bm I}_r}
\newcommand{\gtrgb}{{\bm I}_r}
\newcommand{\ourmethod}[0]{Ours}
\newcommand{\duster}[0]{DUSt3R}
\newcommand{\master}[0]{MASt3R}
\newcommand{\monster}[0]{MonST3R}
\newcommand{\spanner}[0]{Spann3R}
\usepackage{duckuments}

\begin{document}
\begin{figure}
\twocolumn[{%
\renewcommand\twocolumn[1][]{#1}
\maketitle
\centering
\vspace{-1.2em}
\hspace*{-.3in}\includegraphics[width=1\textwidth]{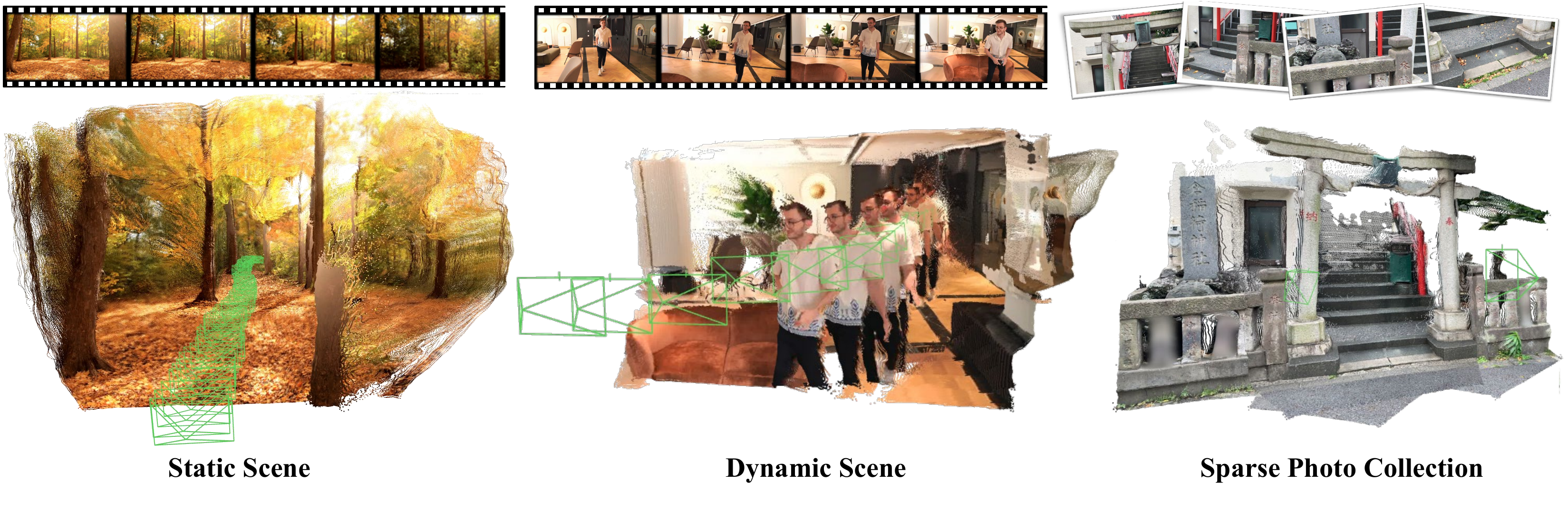}
\vspace{-0.5em}
\caption{\label{fig:teaser}%
\textbf{Continuous 3D Perception.} Given a stream of RGB images as input, our approach enables dense 3D reconstruction in an online, continuous manner, estimating both camera parameters and dense 3D geometry with each incoming frame. Our framework supports various 3D tasks, processes inputs from video sequences and sparse photo collections, and can handle both static and dynamic scenes. 
}
\vspace{1em}
}]

\end{figure}

\vspace{-1em}\begin{abstract}
We present a unified framework capable of solving a broad range of 3D tasks.
Our approach features a stateful recurrent model that continuously updates its state representation with each new observation. Given a stream of images, this evolving state can be used to generate metric-scale pointmaps (per-pixel 3D points) for each new input in an online fashion. These pointmaps reside within a common coordinate system, and can be accumulated into a coherent, dense scene reconstruction that updates as new images arrive.
Our model, called \textbf{CUT3R} (\textbf{C}ontinuous \textbf{U}pdating \textbf{T}ransformer for \textbf{3}D \textbf{R}econstruction), captures rich priors of real-world scenes: not only can it predict accurate pointmaps from image observations,
but it can also infer unseen regions of the scene by probing at virtual, unobserved views.
Our method is simple yet highly flexible, naturally accepting varying length of images that may be either video streams or unordered photo collections, containing both static and dynamic content. We evaluate our method on various 3D/4D tasks and demonstrate competitive or state-of-the-art performance in each. 
Project page: \url{https://cut3r.github.io/}.

\end{abstract}    
\section{Introduction}

Humans are online visual learners. We continuously process streams of visual input, building on what we have learned in the past while learning in the present. Our prior knowledge enables us to interpret the world from minimal information; \textit{e.g.}, upon entering a new restaurant, it only takes a glance to start inferring its layout and atmosphere. But it doesn't stop there---as we accumulate more observations, we continuously refine our mental model of the 3D environment. This ability to reconcile our prior knowledge of the world with a continuous stream of new observations is crucial for functioning effectively in an ever-changing visual world.

Building on these insights, we introduce an online 3D perception framework that unifies three key capabilities: 1) reconstructing 3D scenes from few observations, 2) continuously refining the reconstruction with more observations, and 3) inferring 3D properties of unobserved scene regions. We achieve these capabilities by integrating data-driven priors with a recurrent update mechanism. The learned prior enables our method to address challenges encountered by traditional methods (e.g., dynamic objects, sparse observations, degenerate camera motion), while the ability to continuously update allows it to process new observations online, and improve the reconstruction continuously over time.

\begin{figure}
    \centering
    \includegraphics[width=1.0\linewidth]{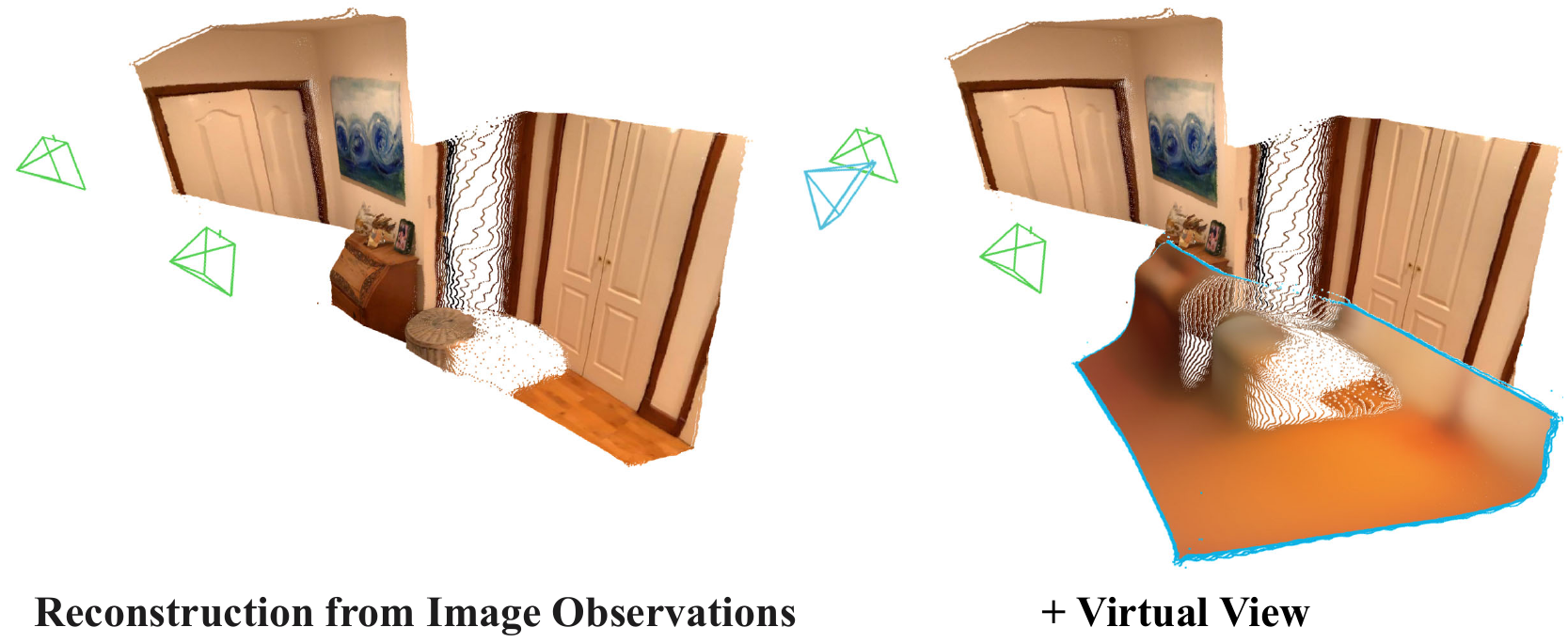}
    \caption{\small \textbf{Querying Unseen Regions.} In addition to reconstructing a scene from images, our method can also infer structure for \emph{unseen} parts of the scene, given a virtual camera query (shown in blue).
    }
    \label{fig:tiny-raymap}
    \vspace{-1.5em}
\end{figure}

Specifically, given an image stream, our recurrent model maintains and incrementally updates a persistent internal state that encodes the scene content. With each new observation, the model simultaneously updates this state and reads from it to predict the current view’s 3D properties, including an estimate of that view's dense 3D geometry (as a pointmap; a 3D point per-pixel in a world coordinate frame) and camera parameters (both intrinsics and extrinsics). Accumulating these pointmaps enables online dense scene reconstruction, as illustrated in Fig. \ref{fig:teaser}. Additionally, our framework supports inferring unobserved parts of the scene: by querying the internal state with a virtual (unseen) view, parameterized as a raymap, we can extract the corresponding pointmap and color for the query view, as depicted in Fig. \ref{fig:tiny-raymap}.

Our framework is designed to be general and flexible, making it well-suited for training on an extensive collection of datasets and adaptable to diverse inference scenarios. During training, we leverage a wide variety of 3D data, including single images, videos, and photo collections with partial or full 3D annotations. These datasets span a broad spectrum of scene types and contexts—static and dynamic, indoor and outdoor, real and synthetic—enabling the model to acquire robust and generalizable priors. During inference,  our recurrent framework naturally accepts varying numbers of images, and supports a wide range of input data settings: from streaming video to unstructured image collections, including wide-baseline or even non-overlapping images. Beyond static scenes, it seamlessly handles videos of dynamic scenes, estimating accurate camera parameters and dense point clouds for moving parts of the scene. 

We evaluate our method on various 3D tasks: monocular and consistent video depth estimation, camera pose estimation, and 3D reconstruction, achieving competitive or state-of-the-art performance in each. We also show that our method can infer previously unseen structures and continuously refine the reconstruction as new observations arrive.

\section{Related Work}

\noindent\textbf{\textit{Tabula rasa} 3D Reconstruction.} Many 3D reconstruction pipelines operate as \textit{tabula rasa}\footnote{tabula rasa (lat.) -- blank slate.}, starting from scratch for each new scene, and relying solely on the observations  currently available.
Notable examples include traditional methods such as Structure from Motion (SfM)~\cite{agarwal2011building,schoenberger2016sfm,snavely2006photo,snavely2008modeling,lowe2004distinctive,hartley2003multiple,triggs2000bundle,agarwal2010bundle,seitz2006comparison} and Simultaneous Localization and Mapping (SLAM)~\cite{cadena2016past,durrant2006simultaneous,davison2007monoslam,mur2015orb,engel2014lsd,klein2007parallel,newcombe2011dtam}, as well as more recent approaches such as Neural Radiance Fields (NeRF)~\cite{mildenhall2021nerf,muller2022instant,chen2022tensorf,fridovich2022plenoxels,wang2021neus} and 3D Gaussian Splatting~\cite{kerbl20233d}. The \textit{tabula rasa} nature of these approaches poses challenges in handling scenarios with sparse observations or ill-posed conditions. In contrast, our method leverages data-driven priors to enable dense 3D reconstruction directly from video sequences or photo collections, eliminating the need for known camera extrinsics or intrinsics.  
The data-driven nature of our method enables it to address challenging cases of degeneracy, make predictions from as few as a single image, and infer structures that are unobserved in the input.

\vspace{0.5em}\noindent\textbf{Learning-Based 3D Reconstruction.} 
Unlike \textit{tabula rasa} reconstruction, many methods integrate data-driven priors into 3D reconstruction. One prominent direction focuses on improving traditional reconstruction pipelines, including replacing hand-crafted components with learning-based alternatives (e.g., substituting conventional feature descriptors~\cite{lowe2004distinctive,bay2008speeded} with learned ones~\cite{sarlin20superglue,detone2018superpoint,sun2021loftr,dusmanu2019d2}), integrating data-driven priors~\cite{teed2021droid,chen2024leap,teed2024deep,yu2024mip,yang2020d3vo,zhu2024nicer,tateno2017cnn} into the systems, or optimizing entire systems end-to-end~\cite{wang2024vggsfm,teed2021droid,tang2018ba,yao2018mvsnet}.
Another substantial body of work aims to predict dense 3D geometry directly from single or paired images.
Extensive research~\cite{li2018megadepth,yang2024depth,godard2019digging,bhat2023zoedepth,wang2024moge,piccinelli2024unidepth,Ranftl2022,Ranftl2021} focuses on estimating monocular depth. While depth maps provide valuable 3D information, converting them into 3D point clouds typically requires camera intrinsics~\cite{piccinelli2024unidepth,hu2024metric3d,depthpro}, with camera extrinsics often left unmodeled.
Notably, DUSt3R~\cite{wang2024dust3r} predicts two pointmaps from an image pair within the same coordinate frame, inherently accounting for both camera intrinsics and extrinsics. However,  DUSt3R is tailored for image pairs and does not inherently support multiple views. While it can be extended to handle multi-view tasks via an additional global alignment process, this process operates offline, is time-consuming, and cannot dynamically update the reconstruction as new observations are added. 
In contrast, our method flexibly handles a varying number of images, predicting pointmaps for each image in a shared coordinate frame as they are received.

\vspace{0.5em}
\noindent\textbf{Continuous Reconstruction Methods.} 
Many traditional and learning-based methods share the ability to continuously predict 3D structure in an online manner. 
Monocular SLAM pipelines~\cite{teed2021droid,zhu2024nicer,engel2014lsd,forster2016svo} recover ego-motion and 3D point clouds in real-time from video sequences but typically require known camera intrinsics.  Our approach is related to learning based methods~\cite{kar2017learning,choy20163d,zhang2022nerfusion,sun2021neuralrecon}, such as 3D-R2N2~\cite{choy20163d}, which utilize recurrent neural network architectures~\cite{elman1990finding,hochreiter1997long,jaegle2021perceiver} for online 3D reconstruction. However, these methods are either object-centric~\cite{choy20163d,kar2017learning,yu2021pixelnerf} or require posed images as input~\cite{kar2017learning,sun2021neuralrecon,sayed2022simplerecon,bozic2021transformerfusion,zhang2022nerfusion}. Spann3R~\cite{spann3r}, developed concurrently with our work, also demonstrates continuous reconstruction capabilities using a spatial memory mechanism.  However, while Spann3R’s memory serves primarily as a cache for observed scenes, our compressed state representation not only captures observed scene content but also enables inferring unobserved structures. Finally, unlike the methods discussed in this paragraph that only operate on static scenes, our method also seamlessly reconstructs dynamic scenes.

\begin{figure*}[t!]
    \centering
    \includegraphics[width=\linewidth]{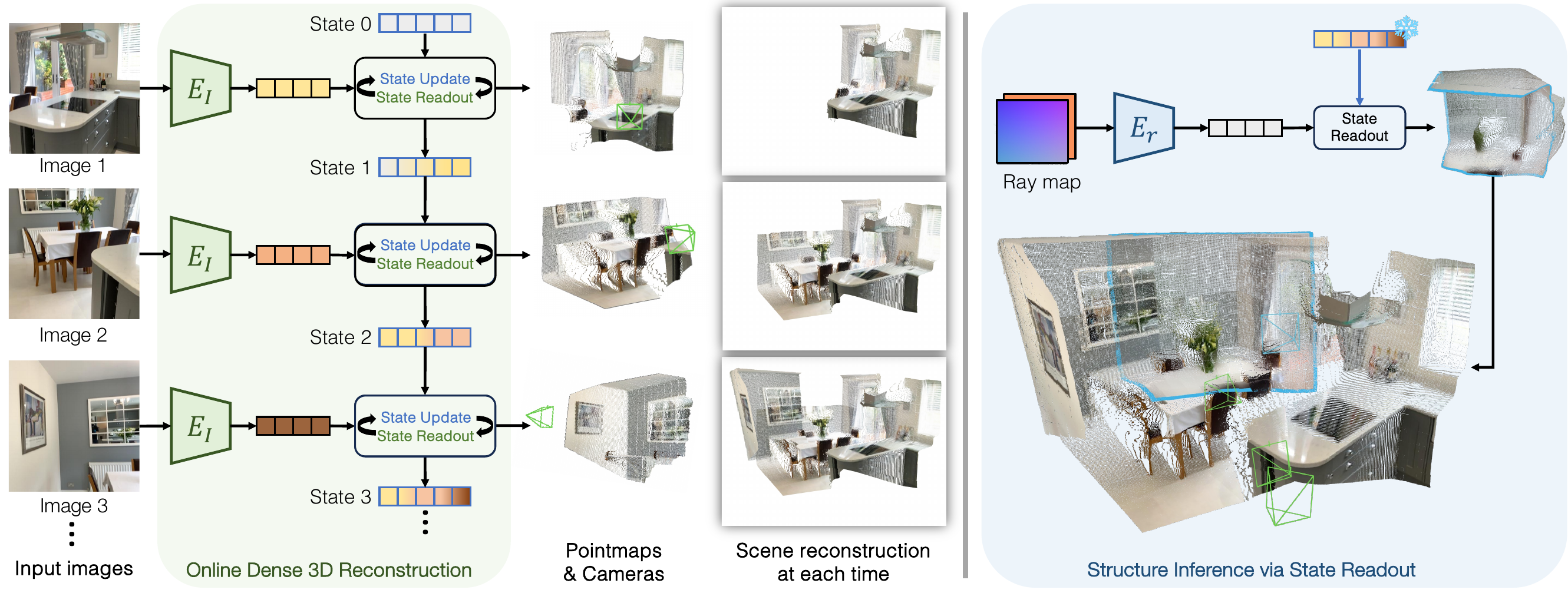}
    \vspace{-.2in}
    \caption{\small \textbf{Method Overview}. Our method performs online dense 3D reconstruction from a stream of images (video frames or a photo collection) by using a persistent state. Each input image is encoded into visual tokens via a shared-weight ViT encoder. These tokens interact with state tokens, where \emph{state update} integrates the current image into the  state, and \emph{state readout} retrieves the past context stored in the state for predictions. Both processes occur simultaneously through two interconnected ViT decoders. Outputs include pointmaps in world and camera frames (only world pointmaps are shown) and the camera-to-world transformation. On the right, we demonstrate our method's ability to predict unseen views: given a query camera (as a raymap), it reads information from the state to predict its corresponding pointmap, even for unobserved regions. For these readouts, we do not update the state. The hallucinated pointmap is highlighted with a blue border.
    }
    \label{fig:overview}
\end{figure*}

\vspace{0.5em}\noindent\textbf{Reconstructing Dynamic Scenes from Monocular Videos.}
Recovering camera parameters and consistent dense geometry from monocular videos of dynamic scenes presents significant challenges for traditional SLAM pipelines. Recent approaches address this by leveraging learned depth priors.
Robust-CVD~\cite{robustcvd} uses deformation splines to align the depth maps, while 
CasualSAM~\cite{casualsam} finetunes a monocular depth network on a single video. These methods involve time-consuming per-video optimization.
Concurrent to our work, MonST3R~\cite{zhang2024monst3r} extends DUSt3R to predict pointmaps for dynamic scenes by finetuning it on dynamic datasets. However, it still follows DUSt3R's pairwise formulation and requires global alignment as post-processing. Another concurrent work, MegaSaM~\cite{li2024megasam}, achieves highly accurate and robust estimation of camera poses and scene structure for casually captured dynamic videos. Unlike our approach, MegaSaM is optimization-based (\textit{i.e.}, not feedforward), uses an explicit 3D state and does not make online predictions.

\vspace{0.5em}\noindent\textbf{3D Scene Priors.} 
Predicting 3D content beyond observed views has long been studied, but it has recently gained significant attention due to advancements in generative modeling. Regression-based few-shot novel view synthesis approaches~\cite{sitzmann2019scene,yu2021pixelnerf,sitzmann2021light,lai2021video} typically only generalize across a class of 3D scenes. With rapid advances in image and video generative models~\cite{rombach2022high,ramesh2021zero,saharia2022photorealistic}, much of the current research in 3D generation focuses on transferring priors from images and videos to 3D~\cite{liu2023zero1to3, melaskyriazi2023realfusion, wu2024reconfusion, poole2022dreamfusion}.
However, most image-based 3D generation methods require camera parameters as input, focus on limited data domains (\textit{e.g.}, object-centric), or require additional 3D distillation processes to extract 3D content. In contrast to these view-centric approaches that generate novel views as a proxy for 3D, our method is geometry-centric:
It directly generates metric-scale pointmaps for a virtual camera query in a scene reconstructed from a set of images, without requiring their camera intrinsics or poses.

\section{Method}

Our approach takes as input a stream of images  without any camera information. The image streams can come from either video or image collections. %
As a new image comes in through the model, it interacts with the latent state representation, which encodes the understanding of the current 3D scene. 
Specifically, the image simultaneously updates the state with new information and retrieves information stored in the state.
Following the state-image interaction, explicit 3D pointmaps and camera poses are extracted for each view.
The state can also be queried with a virtual view to predict its corresponding pointmap, capturing unseen parts of the scene. See Fig.~\ref{fig:overview} for our method overview.

\subsection{State-Input Interaction Mechanism}
\label{sec:state-img}

Our method takes a stream of images as input. For each current image, $\It$, it is first encoded into token representation by a ViT encoder~\cite{dosovitskiy2020image}:

\begin{equation}
    \ft = \mathrm{Encoder}_i(\It).
\end{equation}
We represent the state also as a set of tokens. Prior to seeing any image input, the state tokens are initialized as a set of learnable tokens shared by all scenes. 
The image tokens interact with the state in two ways: they update the state with information from the current image and read the context from the state, incorporating stored past information.
We refer to these interactions as \textit{state-update} and \textit{state-readout}, respectively.
This bidirectional interaction is implemented using two interconnected transformer decoders~\cite{wang2024dust3r,weinzaepfel2022croco}, which jointly operate on both image and state tokens:

\begin{equation}
    [\zts, \fts], \stcur  = \mathrm{Decoders}([\bm z, \ft], \stpre).
    \label{eq:state-img}
\end{equation}
Here $\stpre$ and $\stcur$ represent the state tokens before and after interaction with the image tokens. $\fts$ denotes the image tokens enriched with state information. $\bm z$ is a learnable ``pose token'' prepended to the image tokens,  whose output $\zts$ captures image-level information related to the scene, such as ego motion. %
Within the decoders, the outputs from both sides cross-attend to each other at each decoder block to ensure effective information transfer. 

After this interaction, explicit 3D representation can be extracted from  
$\fts$ and $\zts$. Specifically, we predict two pointmaps with corresponding confidence maps, $(\pmself, \confself)$ and $(\pmcross, \confcross)$. These maps are defined in two coordinate frames: the input image’s own coordinate frame and the world frame, respectively, where the world frame is defined as the coordinate frame of the initial image. Additionally, we
predict the relative transformations between the two coordinate frames, or, the ego motion, $\pose$: 
\begin{align}
    \pmself, \confself &= \headself(\fts) \label{eq:head_self} \\
    \pmcross, \confcross &= \headcross(\fts, \zts) \\
    \pose &= \headpose(\zts),
    \label{eq:head_pose}
\end{align}
where $\headself$ and $\headcross$ are implemented as DPT~\cite{Ranftl2021}, and  $\headpose$ is implemented as an MLP network, respectively. Please see the supplement for details. We extract 6-DoF pose $\pose$ from the pose token $\zts$, the rigid transformation from the current frame to the world. All pointmaps and poses are in metric scale~(i.e., meters).

Although predicting $\pmself$, $\pmcross$, and $\pose$ may seem redundant, 
we found this redundancy simplifies training.  It enables each output to receive direct supervision, and importantly,
it facilitates training on datasets with partial annotations, such as those containing only pose or single-view depth, thereby broadening the range of usable data.

\subsection{Querying the State with Unseen Views}

Leveraging past 3D experiences, humans can envision parts of a scene beyond what is directly observed. We emulate this ability by extending the state-readout operation to predict unseen portions of the scene from a virtual camera view.
Specifically, we use a virtual camera as a query to extract information from the state. The virtual camera's intrinsics and extrinsics are represented as a raymap 
$\raymap$, a 6-channel image encoding the origin and direction of rays at each pixel~\cite{zhang2024raydiffusion, gao2024cat3d, wu2024cat4d}.

Given a query raymap~$\raymap$, we first encode it into token representations $\fr$ using a separate transformer %
$\mathrm{Encoder}_r$:
\begin{equation}
    \fr = \mathrm{Encoder}_r(\raymap).
\end{equation}
Then, the rest of the process aligns largely with what is described in Sec~\ref{sec:state-img}. Specifically, we have
$\fr$ interact with the current state through the same decoder module (i.e., shared weights) as in Eq.~\ref{eq:state-img} to read from the state into 
$\frs$. Note that, unlike in the state-image interaction, the state is not updated here, as the raymap serves solely as a query without introducing new scene content. 
Finally, we apply the same head networks to parse 
$\frs$ into explicit representations, as in Eq.~\ref{eq:head_self}-\ref{eq:head_pose}. Additionally, we introduce another head~$\headcolor$ to decode color information: $ \predrgb = \headcolor(\frs)$ which corresponds to the color of each ray in the raymap $\raymap$.

Querying the scene with raymaps has an interesting analogy to Masked Autoencoders~(MAE)~\cite{he2022masked}. In MAE, completion occurs at the patch level, using the global context of the entire image. Here, completion is performed at the image level, leveraging the global context of the 3D scene captured in the state. %

\subsection{Training Objective}
During training, we provide the model with a sequence of $N$ images (either from a video sequence or an image collection). The raymap mode is enabled only for training data with metric-scale 3D annotations. In these cases, we randomly replace each image with its corresponding raymap at a certain probability, excluding the first view. When the scale of the 3D annotation is unknown, raymap querying is disabled to avoid inconsistencies between the annotation scale and the scene scale represented in the state.  For clarity, we do not differentiate between predictions generated from a raymap or an image in this section to simplify the notation. 
We denote the pointmap predictions of the current training sequence as $\mathcal{X} = \{\pmselfset, \pmcrossset\}$, where
$\pmselfset=\{\pmself\}_{t=1}^N$, $\pmcrossset=\{\pmcross\}_{t=1}^N$ and their corresponding confidence scores as $\mathcal{C}$.

\vspace{-0.5em}
\paragraph{3D regression loss.} Following MASt3R~\cite{leroy2024mast3r}, we apply a confidence-aware regression loss to the pointmaps:
\begin{equation}
\small
    \mathcal{L}_\text{conf} = \sum_{(\hat{\bm x}, c)\in ( \mathcal{\hat{X}}, \mathcal{C})} \left(c\cdot \left\|\frac{\hat{\bm x}}{\hat s} - \frac{\bm x}{s}\right\|_2 - \alpha \log c\right), 
    \vspace{-.5em}
\end{equation}
where $\hat{s}$ and $s$ are scale normalization factors for  $\mathcal{\hat{X}}$ and $\mathcal{X}$, respectively. Similar to MASt3R~\cite{leroy2024mast3r}, when the groundtruth pointmaps are metric, we set $\hat{s}:=s$ to enable the model to learn metric-scale pointmaps. 

\noindent\textbf{Pose loss.} We parameterize the pose $\pose$ as quaternion $\hat{\bm q}_t$ and translation $\hat{\bm \tau}_t$, and minimize the L2 norm between the prediction and groundtruth:
\begin{equation}
\small
    \mathcal{L}_\text{pose} = \sum_{t=1}^N
    \left( \left\|\hat{\bm q}_t - \bm q_t\right\|_2 + \left\| \frac{\hat{\bm \tau}_t}{\hat{s}} - \frac{{\bm \tau}_t}{s} \right\|_2 \right).
\end{equation}

\noindent\textbf{RGB loss.} When the input is raymap, besides the 3D regression loss, we also apply an MSE loss to enforce the predicted pixel colors $\predrgb$ to match the groundtruth: $\mathcal{L}_{rgb} = ||\predrgb - \gtrgb||_2^2$.

\subsection{Training Strategy}
\noindent\textbf{Training Datasets.} We train our method on a diverse set of 32 datasets, covering synthetic and real-world data, static and dynamic scenes, scene-level and object-centric configurations, as well as both indoor and outdoor scenes. Examples of our datasets include CO3Dv2~\cite{reizenstein21co3d}, ARKitScenes~\cite{dehghan2021arkitscenes}, ScanNet++~\cite{yeshwanth2023scannet++}, TartanAir~\cite{tartanair2020iros}, Waymo~\cite{Sun_2020_CVPR}, MegaDepth~\cite{MDLi18}, MapFree~\cite{arnold2022mapfree},
DL3DV~\cite{ling2024dl3dv}, and DynamicStereo~\cite{karaev2023dynamicstereo}. Our flexible formulation allows training on datasets with partial annotations~(\ie only camera parameters like RealEstate10K~\cite{realestate10k}, or only single views, like Synscapes~\cite{wrenninge2018synscapes}). See the supplement for the full list. %

\vspace{.5em}\noindent\textbf{Curriculum Training.}
Our model is trained with a curriculum. The first stage trains the model on 4-view sequences from mainly static datasets.
The second stage incorporates dynamic scene datasets, improving the model's ability to handle moving objects such as humans, along with  datasets with partial annotations, further enhancing its generalization. 
These two stages are trained on 
224$\times$224 images to reduce computational costs, following \duster{}~\cite{wang2024dust3r}.
In the third stage, we train with higher resolution, using varied aspect ratios and setting the maximum side to 512 pixels.
Finally, we freeze the encoder, training only the decoder and heads on longer sequences spanning 4 to 64 views. This stage focuses on enhancing inter-scene reasoning and effectively handling long contexts. %

\vspace{.5em}\noindent\textbf{Implementation Detail.}
We use a ViT-Large model~\cite{dosovitskiy2020image} for the image encoder $\mathrm{Encoder}_i$, initialized with \duster{} encoder pretrained weights, and ViT-Base for the decoders. Both the encoder and decoders operate on 16$\times$16 pixel patches. The state consists of 768 tokens, each with a dimensionality of 768. The raymap encoder~ $\mathrm{Encoder}_r$ is a lightweight encoder with 2 blocks. 
We use Adam-W optimizer~\cite{loshchilov2017decoupled} with an initial learning rate of $1e^{-4}$, applying linear warmup followed by cosine decay. We train our model on eight A100 NVIDIA GPUs each with 80G memory. 
Please refer to the supplement for more details.

\section{Experiments}
We evaluate our method across a range of 3D tasks, including single and video depth estimation~(Sec.~\ref{sec:exp_depth}), camera pose estimation~(Sec.~\ref{sec:exp_pose}), and 3D reconstruction~(Sec.~\ref{sec:exp_recon}). 

\vspace{0.5em}\noindent\textbf{Baselines.}
Our primary set of baselines are \duster{}~\cite{wang2024dust3r}, \master{}~\cite{leroy2024mast3r}, \spanner{}~\cite{spann3r}, and \monster{}~\cite{zhang2024monst3r}, where the latter two are concurrent works. \monster{} finetunes \duster{} on dynamic datasets to handle dynamic scenes, while \spanner{} extends \duster{} to support varying number of images via additional spatial memory and operates online, similar to our method. \duster{}, \master{}, and \monster{} can only take a pair of views as input, and require an extra global alignment~(GA) stage to consolidate the pairwise predictions. Both \master{} and our method predict metric pointmaps, whereas others predict relative pointmaps.

\subsection{Monocular and Video Depth Estimation}
\label{sec:exp_depth}

\paragraph{Mono-Depth Estimation.} Following MonST3R~\cite{zhang2024monst3r}, 
we evaluate monocular depth estimation on KITTI~\cite{kitti}, Sintel~\cite{sintel}, Bonn~\cite{bonn} and NYU-v2~\cite{nyuv2} datasets covering dynamic and static, indoor and outdoor scenes. These datasets are excluded from training, enabling zero-shot performance evaluation across domains.
We use absolute relative error (Abs Rel) and $\delta < 1.25$ (percentage of predicted depths within a $1.25$-factor of true depth) as metrics, with per-frame median scaling per \duster{}~\cite{wang2024dust3r}. Results in Tab.~\ref{tab:single_frame_depth} show our method achieves state-of-the-art or competitive performance, leading on Bonn and and NYU-v2 and ranking second on KITTI.

\begin{table}[H]
\centering
\renewcommand{\arraystretch}{0.95}
\renewcommand{\tabcolsep}{2.pt}
\resizebox{\linewidth}{!}{
\begin{tabular}{@{}lcc|cc|cc|cc@{}}
\toprule
  & \multicolumn{2}{c}{\textbf{Sintel}} & \multicolumn{2}{c}{\textbf{Bonn}} & \multicolumn{2}{c}{\textbf{KITTI}} & \multicolumn{2}{c}{\textbf{NYU-v2}} \\ 

\cmidrule(lr){2-3} \cmidrule(lr){4-5} \cmidrule(lr){6-7} \cmidrule(lr){8-9}
 {\textbf{Method}} & {\footnotesize Abs Rel $\downarrow$} & {\footnotesize $\delta$\textless{}$1.25\uparrow$} & {\footnotesize Abs Rel $\downarrow$} & {\footnotesize $\delta$\textless{}$1.25\uparrow$} & {\footnotesize Abs Rel $\downarrow$} & {\footnotesize $\delta$\textless{}$1.25\uparrow$} & {\footnotesize Abs Rel $\downarrow$} & {\footnotesize $\delta$\textless{}$1.25\uparrow$} \\ 
\midrule
DUSt3R & 0.424 & \underline{58.7} & 0.141 & 82.5 & 0.112 & 86.3 & \textbf{0.080} & \underline{90.7} \\ 
MASt3R & \textbf{0.340} & \textbf{60.4} & {0.142} & {82.0} & \textbf{0.079} & \textbf{94.7} & {0.129} & 84.9 \\ 
MonST3R & \underline{0.358} & 54.8 & \underline{0.076} & \underline{93.9} & {0.100} & {89.3} & 0.102 & 88.0 \\ 
Spann3R & {0.470} & 53.9 & {0.118} & {85.9} & {0.128} & {84.6} & 0.122 & 84.9 \\ 
\ourmethod{} & {0.428} & {55.4} & \textbf{0.063} & \bf{96.2} & \underline{0.092} & \underline{91.3} & \underline{0.086} & \textbf{90.9} \\ 
\bottomrule
\end{tabular}
}
\caption{\small \textbf{Single-frame Depth Evaluation.} We report the performance on Sintel, Bonn, KITTI, and NYU-v2 (static) datasets. }
\label{tab:single_frame_depth}
\vspace{-.1in}

\end{table}

\begin{table*}[t]
\centering
\renewcommand{\arraystretch}{1.02}
\renewcommand{\tabcolsep}{1.5pt}
\resizebox{0.85\textwidth}{!}{
\begin{tabular}{@{}llcc>{\centering\arraybackslash}p{1.5cm}>{\centering\arraybackslash}p{1.5cm}|>{\centering\arraybackslash}p{1.5cm}>{\centering\arraybackslash}p{1.5cm}|>{\centering\arraybackslash}p{1.5cm}>{\centering\arraybackslash}p{1.5cm}|>{\centering\arraybackslash}p{1.2cm}@{}}
\toprule
 &  &  &  & \multicolumn{2}{c}{\textbf{Sintel}} & \multicolumn{2}{c}{\textbf{BONN}} & \multicolumn{2}{c}{\textbf{KITTI}} & \\ 
\cmidrule(lr){5-6} \cmidrule(lr){7-8} \cmidrule(lr){9-10}
\textbf{Alignment} & \textbf{Method} & \textbf{Optim.} & \textbf{Onl.\ } & {Abs Rel $\downarrow$} & {$\delta$\textless $1.25\uparrow$} & {Abs Rel $\downarrow$} & {$\delta$\textless $1.25\uparrow$} & {Abs Rel $\downarrow$} & {$\delta$ \textless $1.25\uparrow$} & \textbf{FPS} \\ 
\midrule

\multirow{6}{*}{\begin{minipage}{3cm}Per-sequence scale\end{minipage}} & DUSt3R-GA~\cite{wang2024dust3r} &  \checkmark & & 0.656 & {45.2} & {0.155} & {83.3} & \underline{0.144} & \underline{81.3} & 0.76 \\
& MASt3R-GA~\cite{leroy2024mast3r} &   \checkmark & & 0.641 & {43.9} & {0.252} & {70.1} & {0.183} & {74.5} & 0.31 \\

& MonST3R-GA~\cite{zhang2024monst3r} &  \checkmark &  & \textbf{0.378} & \textbf{55.8} & \textbf{0.067} & \textbf{96.3} & {0.168} & {74.4} & 0.35 \\
& Spann3R~\cite{spann3r} &  & \checkmark & 0.622 & {42.6} & {0.144} & {81.3} & {0.198} & {73.7} &13.55 \\

& \ourmethod &  & \checkmark & \underline{0.421}  & \underline{47.9} & \underline{0.078} & \underline{93.7} & \textbf{0.118} & \textbf{88.1} &  16.58 \\

\midrule
 \multirow{2}{*}{\begin{minipage}{3cm}Metric scale\end{minipage}} & MASt3R-GA~\cite{leroy2024mast3r} & \checkmark &  & \textbf{1.022}  & 14.3 & 0.272 & 70.6 & 0.467 & 15.2  & 0.31 \\  
& \ourmethod &  & \checkmark & {1.029} & \textbf{23.8} & \textbf{0.103} & \textbf{88.5}& \textbf{0.122} & \textbf{85.5}  &  16.58 \\
\bottomrule
\end{tabular}
}
\vspace{-.05in}
\caption{\small{
\textbf{Video Depth Evaluation}. We report scale-invariant depth and metric depth accuracy on Sintel, Bonn, and KITTI datasets. Methods requiring global alignment are marked ``GA'', while ``Optim.'' and ``Onl.'' indicate optimization-based and online methods, respectively. We also report the FPS on KITTI dataset using 512$\times$ 144 image resolution for all methods on an A100 GPU, except \spanner{} which only supports 224$\times$224 inputs.
We present a subset of baselines here; please refer to the supplementary material for full comparisons. 
}
}
\label{tab:video_depth}
\end{table*}

\begin{table*}[t]
\centering
\footnotesize
\renewcommand{\arraystretch}{1.}
\renewcommand{\tabcolsep}{2.5pt}
\resizebox{0.9\textwidth}{!}{
\begin{tabular}{@{}clccccc|ccc|ccc@{}}
\toprule
& & & & \multicolumn{3}{c}{\textbf{Sintel}} & \multicolumn{3}{c}{\textbf{TUM-dynamics}} & \multicolumn{3}{c}{\textbf{ScanNet}} \\ 
\cmidrule(lr){5-7} \cmidrule(lr){8-10} \cmidrule(lr){11-13}
{} & {\textbf{Method}}  &  \textbf{Optim.} & \textbf{Onl.} & {ATE $\downarrow$} & {RPE trans $\downarrow$} & {RPE rot $\downarrow$} & {ATE $\downarrow$} & {RPE trans $\downarrow$} & {RPE rot $\downarrow$} & {ATE $\downarrow$} & {RPE trans $\downarrow$} & {RPE rot $\downarrow$} \\ 
\midrule
& Particle-SfM~\cite{zhao2022particlesfm} & \checkmark & & \underline{0.129} & {0.031} & \bf{0.535} & - & - & - & 0.136 & 0.023 & 0.836 \\ 
& Robust-CVD~\cite{robustcvd} & \checkmark & & 0.360 & 0.154 & 3.443 & 0.153 & 0.026 & 3.528 & 0.227 & 0.064 & 7.374 \\ 
 & CasualSAM~\cite{casualsam} & \checkmark & & 0.141 & \textbf{0.035} & \underline{0.615} & \underline{0.071} & \textbf{0.010} & 1.712 & 0.158 & 0.034 & 1.618 \\ 
 & DUSt3R-GA~\cite{wang2024dust3r} & \checkmark & & 0.417 & 0.250 & 5.796 & 0.083 & 0.017 & 3.567 & {0.081} & 0.028 & 0.784 \\  
{ }& MASt3R-GA~\cite{leroy2024mast3r} & \checkmark & & {{0.185}} & {0.060} & {1.496} & {\bf{0.038}} & {\underline{0.012}} & {\bf{0.448}} & {\underline{0.078}} & {\underline{0.020}} & {\bf {0.475}} \\ 
& MonST3R-GA~\cite{zhang2024monst3r}  & \checkmark & & \bf {{0.111}} & \underline{0.044} & {0.869} & {{0.098}} & {{0.019}} & {\underline{0.935}} & {\bf{0.077}} & {\bf{0.018}} & {\underline{0.529}} \\ 
\midrule
&  \duster{}~\cite{wang2024dust3r} & &\checkmark  & {\underline{0.290}} & {0.132} & {7.869} & {{0.140}} & {{0.106}} & {{3.286}} & {{0.246}} & {{0.108}} & {{8.210}} \\ 

& Spann3R~\cite{spann3r} & &\checkmark  & {{0.329}} & \underline{0.110} & \underline{4.471} & {\underline{0.056}} & {\underline{0.021}} & {\underline{0.591}} & {\bf{0.096}} & {\underline{0.023}} & {\underline{0.661}} \\ 

& Ours &  & \checkmark & \bf 0.213 & \bf 0.066 & \bf 0.621 & \textbf{0.046} & \bf 0.015 & \bf 0.473 & \underline{0.099} & \textbf{0.022} & \textbf{0.600}\\ 
\bottomrule
\end{tabular}
}
\vspace{-.1in}
\caption{\textbf{Evaluation on Camera Pose Estimation} on Sintel~\cite{sintel}, TUM-dynamic~\cite{tum-dynamics}, and ScanNet~\cite{dai2017scannet} datasets.
Our method achieves the best overall performance among all online methods.
}
\label{tab:video_pose}
\end{table*}

\vspace{0.5em}\noindent\textbf{Video Depth Estimation.}
Video depth estimation evaluates per-frame depth quality and inter-frame depth consistency by aligning predicted depth maps to ground truth using a per-sequence scale. 
For metric pointmap methods like ours and \master{}, we also report results without alignment. 
Comparisons for both methods are presented in Tab.~\ref{tab:video_depth}.

Under per-sequence scale alignment,
our method consistently outperforms \duster{}~\cite{wang2024dust3r} and \master{}~\cite{leroy2024mast3r}. The global alignment they use assumes that the scene is static, and enforcing multi-view consistency can only improve the reconstruction of static regions but may impair the reconstruction of moving objects. In contrast, our method leverages implicit alignment through the state, making it adaptable to both static and dynamic scenes while remaining fully online. Our method also significantly outperforms the other online method, \spanner{}~\cite{spann3r}, which is based on spatial memory designed with a static scene assumption and trained only on static 3D scenes.
\monster{}~\cite{zhang2024monst3r} achieves state-of-the-art performance, but it relies on an additional global alignment stage that incorporates extra input like optical flow in its optimization. In comparison, our method performs comparably, or even better on the KITTI dataset, while remaining online and achieving nearly 50$\times$ speedup. 
In the metric-scale setting, our method also significantly outperforms \master{} for most metrics.

\subsection{Camera Pose Estimation}
\label{sec:exp_pose}

Following ~\monster{}~\cite{zhang2024monst3r}, we evaluate camera pose estimation accuracy on Sintel~\cite{sintel}, TUM dynamics~\cite{tum-dynamics}, and ScanNet~\cite{dai2017scannet} datasets. Note that both Sintel and TUM-dynamics contain major dynamic objects, making it challenging for traditional SfM and SLAM systems.
We report Absolute Translation Error (ATE), Relative Translation Error (RPE trans), and Relative Rotation Error (RPE rot) after Sim(3) alignment with the ground truth, as in \cite{chen2024leap,zhao2022particlesfm,zhang2024monst3r}. Unlike most visual odometry methods~\cite{chen2024leap,he2016deep,teed2021droid}, our method does not require any camera calibration. 
We compare to baselines that share this feature.
Most prior approaches do so through test-time optimization, as seen in Robust-CVD~\cite{robustcvd} and CasualSAM~\cite{casualsam}, which jointly estimate camera parameters and dense depth maps per sequence. 

The results are presented in Tab.~\ref{tab:video_pose}. We separately highlight the leading approaches for methods that require additional optimization and those that do not (i.e., online). 
For the online category, we additionally include \duster{}~\cite{wang2024dust3r} where we align all video frames with first frame, without using GA. Although a gap persists between optimization-based and online methods, our approach achieves the best overall performance among online methods, particularly in dynamic scenes.

\begin{figure*}
    \centering
    \includegraphics[width=\linewidth]{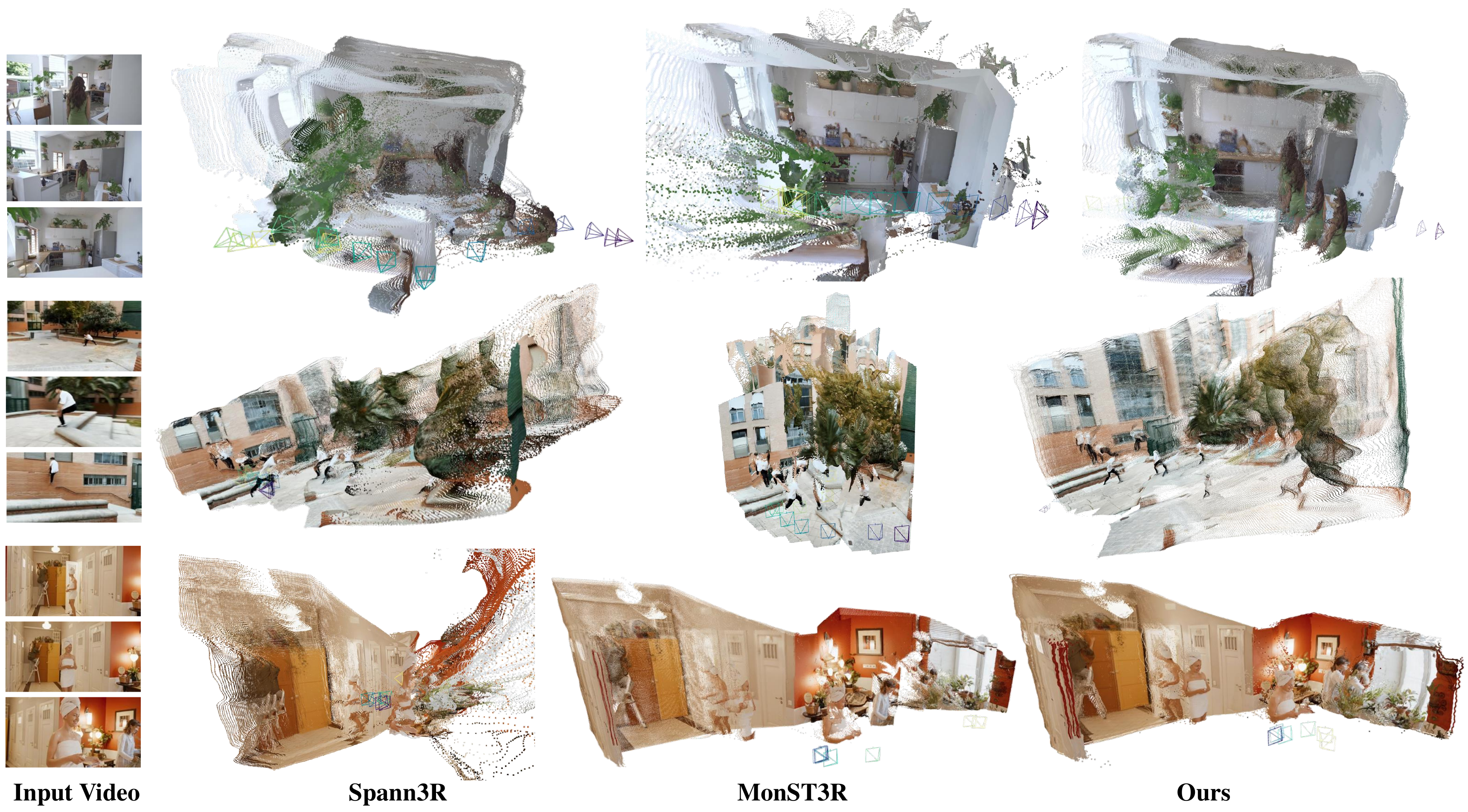}
    \caption{\textbf{Qualitative Results on In-the-wild Internet Videos.} We compare our method with concurrent works \spanner{}~\cite{spann3r} and \monster{}~\cite{zhang2024monst3r}. Our method achieves the best qualitative results. 
    }
    \label{fig:comparison}
\end{figure*}

\begin{table*}[t]
  \centering
  \footnotesize
  \setlength{\tabcolsep}{0.3em}
    \begin{tabularx}{\textwidth}{l c c >{\centering\arraybackslash}X >{\centering\arraybackslash}X >{\centering\arraybackslash}X >{\centering\arraybackslash}X >{\centering\arraybackslash}X >{\centering\arraybackslash}X >{\centering\arraybackslash}X >{\centering\arraybackslash}X >
    {\centering\arraybackslash}X >{\centering\arraybackslash}X >
    {\centering\arraybackslash}X >
    {\centering\arraybackslash}X >{\centering\arraybackslash}X}
      \toprule

           & &  & \multicolumn{6}{c}{\textbf{7 scenes~\cite{shotton2013scene} }} & \multicolumn{6}{c}{\textbf{NRGBD~\cite{azinovic2022neural}}} & \\

      \cmidrule(lr){4-9} \cmidrule(lr){10-15}
         & & & \multicolumn{2}{c}{{Acc}$\downarrow$} & \multicolumn{2}{c}{{Comp}$\downarrow$} & \multicolumn{2}{c}{{NC}$\uparrow$} & \multicolumn{2}{c}{{Acc}$\downarrow$} & \multicolumn{2}{c}{{Comp}$\downarrow$} & \multicolumn{2}{c}{{NC}$\uparrow$} & \\
      \cmidrule(lr){4-5} \cmidrule(lr){6-7} \cmidrule(lr){8-9} \cmidrule(lr){10-11} \cmidrule(lr){12-13} \cmidrule(lr){14-15}
         \textbf{Method} & \textbf{Optim.} & \textbf{Onl.} & {Mean} & {Med.} & {Mean} & {Med.} & {Mean} & {Med.} & {Mean} & {Med.} & {Mean} & {Med.} & {Mean} & {Med.} & \textbf{FPS} \\
      \midrule
      DUSt3R-GA~\cite{wang2024dust3r} &\checkmark &
        & \underline{0.146}	& \underline{0.077}	& {0.181}	& \underline{0.067}	& \bf 0.736	& \bf 0.839	&  0.144	& \bf{0.019}	& 0.154	& \bf{0.018}	& \bf 0.870	& \bf 0.982	& 0.68 \\
        MASt3R-GA~\cite{leroy2024mast3r} &\checkmark &
        & 0.185 & 0.081 & \underline{0.180} & 0.069 & 0.701	& 0.792	& \bf 0.085 & 0.033	& \bf 0.063	& 0.028	& 0.794 & 0.928	& 0.34 \\
        \monster{}-GA~\cite{leroy2024mast3r} &\checkmark &
        & 0.248 & 0.185 & 0.266 & 0.167 & 0.672 & 0.759 & 0.272 & 0.114 & 0.287 & 0.110 & 0.758 & 0.843 & 0.39 \\
      Spann3R~\cite{spann3r} & & \checkmark
        & 0.298 & 0.226 & 0.205 &  0.112 & {0.650} & {0.730}   & {0.416}  & 0.323  &  {0.417}  & {0.285}  & 0.684  & 0.789  & 12.97 \\
        \textbf{\ourmethod} & & \checkmark
       & \bf 0.126 & \bf{0.047} & \bf 0.154 & \bf{0.031} & \underline{0.727} & \underline{0.834} & \underline{0.099} &  \underline{0.031} & \underline{0.076} & \underline{0.026} & \underline{0.837} & \underline{0.971} & 17.00  \\
      \bottomrule
    \end{tabularx}%
    \caption{\textbf{3D reconstruction comparison on 7-Scenes~\cite{shotton2013scene} and NRGBD~\cite{azinovic2022neural} datasets.} While operating online, our method achieves competitive performance, on par with and even surpassing offline methods that employ global alignment.} 
    \label{tab:3d_recon}
\end{table*}

\subsection{3D Reconstruction}
\label{sec:exp_recon}
We evaluate scene-level reconstruction on the 7-scenes~\cite{shotton2013scene} and NRGBD~\cite{azinovic2022neural} datasets using accuracy (Acc), completion (Comp), and normal consistency (NC) metrics, as in prior works~\cite{spann3r, azinovic2022neural, zhu2022nice, wang2023co, wang2024dust3r}.
To assess performance on image collections with minimal or no overlap, we evaluate using sparsely sampled images: 3 to 5 frames per scene for the 7-Scenes dataset and 2 to 4 frames per scene for the NRGBD dataset.
The results are presented in Tab.~\ref{tab:3d_recon}. Our method significantly outperforms the other online approach \spanner{}~\cite{spann3r}, and achieves comparable or sometimes better results than the top optimization-based method, \duster{}-GA, while operating online at 25$\times$ the speed. This highlights our method’s effectiveness with sparse image collections.

\vspace{0.5em}
\noindent\textbf{Qualitative Results.} 
We compare the reconstruction quality of our method with \spanner{}~\cite{spann3r} and \monster{}~\cite{zhang2024monst3r} on in the wild Internet videos in Fig.~\ref{fig:comparison}. \spanner{}~\cite{spann3r} is neither designed nor trained on dynamic scenes, making it less effective at handling moving objects, such as humans.
\monster{}~\cite{zhang2024monst3r} is finetuned on dynamic scenes, potentially overfitting and degrading performance on static 3D scenes. %
In contrast, our method operates online and achieves state-of-the-art performance across both static and dynamic scenes.

\subsection{Analysis}
\vspace{0.5em}\noindent\textbf{State Update Analysis.}
Our model continuously updates its state representation as new data arrives, relying solely on past and current observations without knowledge of future inputs. As more observations accumulate, the state should be able to refine its understanding of the 3D world, leading to improved predictions.
We demonstrate this capability of our method in Tab.~\ref{tab:revisiting}. Using the same experimental setup as in Sec.~\ref{sec:exp_recon}, we introduce an additional version of our approach called ``revisiting'': we first run our method online to obtain the final state that has seen all images, then we freeze this state and use it to process the same set of images again to generate predictions.

\begin{figure}
    \centering
    \includegraphics[width=\linewidth]{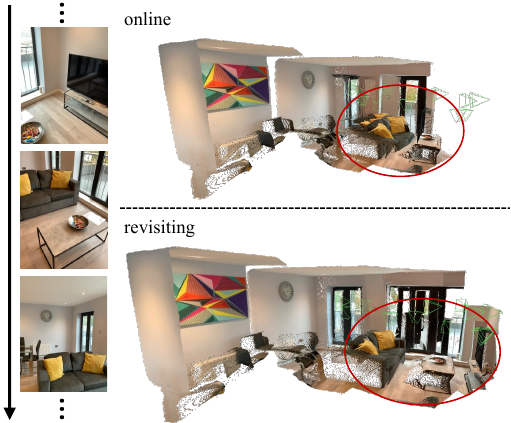}
    \caption{\small \textbf{State Update Analysis.} Compared to online, revisiting incorporates global context which improves overall reconstruction results, especially in the highlighted regions.}
    \vspace{-1em}
    \label{fig:enter-label}
\end{figure}

This setup differs from the online setup by allowing the state to see the full context of the scene during the first run. As shown in Tab.~\ref{tab:revisiting}, revisiting improves performance compared to the online version, especially for accuracy.
This verifies that the state representation effectively updates with additional observations. See Fig.~\ref{fig:enter-label} for a qualitative example.
\begin{table}[t]
  \centering
  \footnotesize
  \setlength{\tabcolsep}{0.3em}
    \begin{tabularx}{\linewidth}{
        l %
        >{\centering\arraybackslash}X %
        >{\centering\arraybackslash}X %
        >{\centering\arraybackslash}X %
        >{\centering\arraybackslash}X %
        >{\centering\arraybackslash}X %
        >{\centering\arraybackslash}X %
    }
      \toprule
       & \multicolumn{3}{c}{\textbf{7-Scenes}} & \multicolumn{3}{c}{\textbf{NRGBD}} \\
      \cmidrule(lr){2-4} \cmidrule(lr){5-7}
       \textbf{Method}& {Acc}$\downarrow$ & {Comp}$\downarrow$ & {NC}$\uparrow$ & {Acc}$\downarrow$ & {Comp}$\downarrow$ & {NC}$\uparrow$ \\
      \midrule
      DUSt3R-GA~\cite{wang2024dust3r} & {0.146} & {0.181} & \textbf{0.736} & {0.144} & {0.154} & \textbf{0.870} \\
      \textbf{\ourmethod} & \underline{0.126} & \underline{0.154} & {0.727} & \underline{0.099} & \textbf{0.076} & {0.837} \\
      \textbf{\ourmethod{} Revisit} & \textbf{0.113} & \textbf{0.107} & \underline{0.732} & \textbf{0.094} & \textbf{0.076} & \underline{0.844} \\
      \bottomrule
    \end{tabularx}%
    \vspace{-5pt}
    \caption{\small
    \textbf{State Update Analysis} on 7-Scenes~\cite{shotton2013scene} and NRGBD~\cite{azinovic2022neural} datasets. 
    }
    \vspace{-1em}
    \label{tab:revisiting}
\end{table}

\begin{figure}
    \centering
    \includegraphics[width=\linewidth]{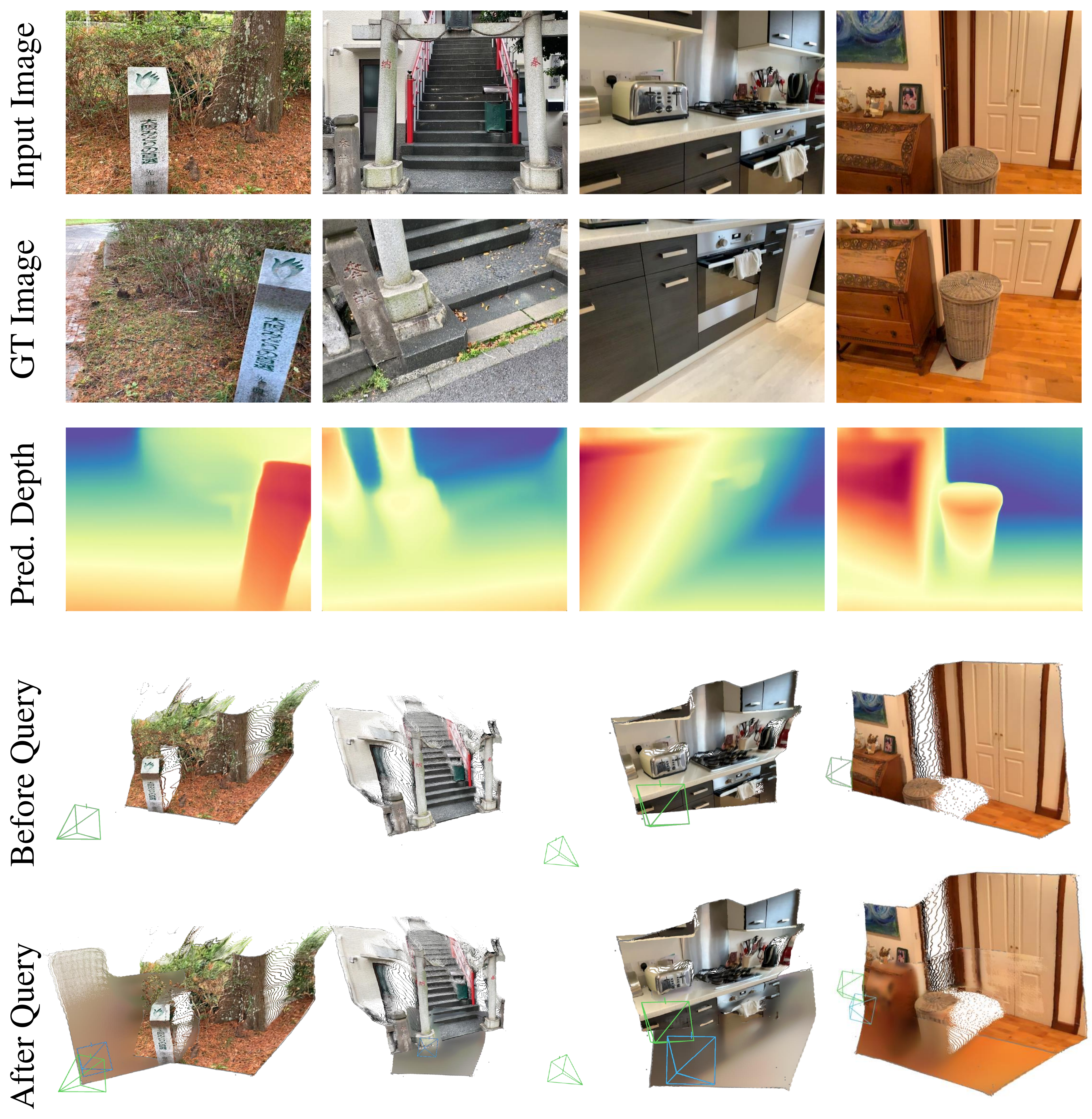}
    \caption{\small \textbf{Inferring New Structure via. State Readout}. From top to bottom: the input image; the ground truth (GT) image, used to query the state via its camera parameters (note: GT image is not given to the model); the depth map from the predicted pointmap; 
    the pointmap prediction of the input image alone; and the pointmap combined with the predicted pointmap in a shared coordinate frame.
    }
    \vspace{-1em}
    \label{fig:structure_generation}
\end{figure}

\vspace{0.5em}
\noindent\textbf{Inferring Unseen Regions via State Readout.} 
To the best of our knowledge, our method is the first to enable the inference of unseen structures in metric scale for general scenes, supporting both single and multiple views, without requiring camera intrinsics or poses for the input images.
We show qualitative results of the generated structures in Fig.~\ref{fig:structure_generation}. 
For this experiment, we use the validation set of the MapFree~\cite{arnold2022mapfree} and ARKitScenes datasets, both with metric camera pose annotations. Importantly, these scenes are not seen by our model during training. 
In each example, we input a single image to our model and then query the state using a raymap of the ground truth image~(unseen by the state). The model is expected to generate pointmaps that align with the ground truth. 
While the predictions may lack some high-frequency details -- owing to the deterministic nature of our approach -- they accurately follows the
viewpoint transformation, 
even with significant viewpoint changes between the input and ground truth. In addition, our method generates new structures beyond what what is observed in the input, such as the bushes in the first example, the ground in the second, the oven in the third, and the stool in the last.  This demonstrates that our method captures generalized 3D scene priors.

\section{Conclusion}
In this paper, we propose an online 3D perception model with a continuously updating, persistent state. Given an image stream, our model simultaneously performs state-update (which updates the state) and state-readout (which retrieves information from the state) for each observation. The output at each step includes camera parameters and pointmaps in the world frame, which accumulate into a dense reconstruction of the scene over time. This simple formulation is general yet powerful enough to solve a number of 3D/4D tasks, handling both videos and photo collections, and processing both static and dynamic scenes. 
The generalized 3D scene priors captured by our method enable the inference of new structures unobserved in the input views by probing the state with a raymap. Experimental results on extensive 3D/4D tasks verify the effectiveness of our method.

\vspace{-1em}
\paragraph{Limitations.} 
As with many online methods, our approach may eventually drift over very long sequences due to the absence of global alignment. Extending our work with explicit or implicit global alignment is an interesting future direction. Additionally, since our structure generation is performed via a deterministic rather than a generative approach, it can produce blurry results – especially when extrapolating viewpoints too far from the provided views  – a common issue with regression-based methods.  Incorporating a generative formulation could address this limitation. Finally, training recurrent networks can be time-consuming. We leave these fundamental directions for future work.

\paragraph{Acknowledgements.} We would like to thank Noah Snavely, Haiwen Feng, Chung Min Kim,  Justin Kerr, Songwei Ge, Chenfeng Xu, Letian Fu, and Ren Wang for helpful discussions. We especially thank Noah Snavely for his guidance and support.
This project is supported in part by DARPA No.~HR001123C0021, IARPA DOI/IBC No.~140D0423C0035, NSF:CNS-2235013, Bakar Fellows, ONR, MURI, TRI and BAIR Sponsors. The views and conclusions contained herein are those of the authors and do not represent the official policies or endorsements of these institutions.

{
    \small
    \bibliographystyle{ieeenat_fullname}
    \bibliography{ref}
}
\clearpage
\appendix
\section*{Appendix}

\section{Training Datasets}
We trained our model on 32 datasets that covers a diverse range of scene types, including static and dynamic environments, as well as indoor, outdoor, and object-centric scenarios. A complete list of these datasets is provided in Tab.~\ref{tab:dataset_info}. 

\begin{table*}[ht]
\centering
\resizebox{\textwidth}{!}{%
\begin{tabular}{l c c c c c c}
\toprule
\textbf{Dataset Name} & \textbf{Scene Type} & \textbf{Metric?} & \textbf{Real?}  & \textbf{Dynamic?} & \textbf{Camera only?} & \textbf{Single View?} \\
\midrule
ARKitScenes~\cite{dehghan2021arkitscenes} & Indoor & Yes & Real & Static & No & No \\
ARKitScenes-HighRes~\cite{dehghan2021arkitscenes} & Indoor & Yes & Real & Static & No & No \\
ScanNet~\cite{dai2017scannet} & Indoor & Yes & Real & Static & No & No \\
ScanNet++~\cite{yeshwanth2023scannet++} & Indoor & Yes & Real & Static & No & No \\
TartanAir~\cite{tartanair2020iros} & Mixed & Yes & Synthetic & Dynamic & No & No \\
Waymo~\cite{Sun_2020_CVPR} & Outdoor & Yes & Real & Dynamic & No & No \\
MapFree~\cite{arnold2022mapfree} & Outdoor & Yes & Real & Static & No & No \\
BlendedMVS~\cite{yao2020blendedmvs} & Mixed & No & Synthetic & Static & No & No \\
HyperSim~\cite{hypersim} & Indoor & Yes & Synthetic & Static & No & No \\
MegaDepth~\cite{li2018megadepth} & Outdoor & No & Real & Static & No & No \\
Unreal4K~\cite{tosi2021smd} & Mixed & Yes & Synthetic & Static & No & No \\
DL3DV~\cite{ling2024dl3dv} & Mixed & No & Real & Static & No & No \\
CO3Dv2~\cite{reizenstein21co3d} & Object-Centric & No & Real & Static & No & No \\
WildRGBD~\cite{xia2024wildrgbd} & Object-Centric & Yes & Real & Static & No & No \\
VirtualKITTI2~\cite{cabon2020vkitti2} & Outdoor & Yes & Synthetic & Dynamic & No & No \\
Matterport3D~\cite{chang2017matterport3d} & Indoor & Yes & Real & Static& No & No \\
BEDLAM~\cite{black2023bedlam} & Mixed & Yes & Synthetic & Dynamic & No & No \\
Dynamic Replica~\cite{karaev2023dynamicstereo} & Indoor & Yes & Synthetic & Dynamic & No & No \\
PointOdyssey~\cite{zheng2023pointodyssey} & Mixed & Yes & Synthetic & Dynamic & No & No \\
Spring~\cite{mehl2023spring} & Mixed & Yes & Synthetic & Dynamic & No & No \\
MVS-Synth~\cite{DeepMVS} & Outdoor & Yes & Synthetic & Dynamic & No & No \\
UASOL~\cite{bauer2019uasol} & Outdoor & Yes & Real & Static & No & No \\
OmniObject3D~\cite{wu2023omniobject3d} & Object-Centric & Yes & Synthetic & Static & No & No \\
RealEstate10K~\cite{realestate10k} & Indoor & No & Real & Static & Yes & No \\
MVImgNet~\cite{yu2023mvimgnet} & Object-Centric & No & Real & Static & Yes & No \\
CoP3D~\cite{sinha2023cop3d} & Object-Centric & No & Real & Dynamic & Yes & No \\
EDEN~\cite{le2021eden} & Outdoor & Yes & Synthetic & Static & No & Yes \\
IRS~\cite{wang2021irs} & Indoor & Yes & Synthetic & Static & No & Yes \\
Synscapes~\cite{wrenninge2018synscapes} & Outdoor & Yes & Synthetic & Dynamic & No & Yes \\
3D Ken Burns~\cite{3dkb} & Mixed & No & Synthetic & Static & No & Yes \\
SmartPortraits~\cite{kornilova2022smartportraits} & Indoor & Yes & Real & Dynamic & No & Yes \\
UrbanSyn~\cite{gómez2023urbansyn} & Outdoor & Yes & Synthetic & Dynamic & No & Yes \\
HOI4D~\cite{liu2022hoi4d} & Indoor & Yes & Real & Dynamic & No & Yes \\
\bottomrule
\end{tabular}%
}
\caption{\textbf{Training Datasets.} We provide more details of our training datasets. We classify a dataset as dynamic if annotations exist for moving objects like humans. If there is only camera parameters (intrinsics and extrinsics) available, we mark them as ``camera only''. If the dataset only contains depth and intrinsics for single views, we mark them as ``single view''.}
\label{tab:dataset_info}
\end{table*}

The original MapFree~\cite{arnold2022mapfree} and DL3DV~\cite{ling2024dl3dv} datasets do not include dense depth maps. We performed multi-view stereo~(MVS) reconstruction~\cite{schoenberger2016sfm} using the provided camera parameters to generate dense depth maps. This results in complete annotations for these datasets for training. RealEstate10K~\cite{realestate10k}, CoP3D~\cite{sinha2023cop3d}, and MVImgNet~\cite{yu2023mvimgnet} also do not provide dense depth maps. For these three datasets, we only use the provided camera parameters to supervise the camera prediction. For RealEstate10K, we only include a subset of 2325 training scenes for training.

EDEN~\cite{le2021eden}, IRS~\cite{wang2021irs}, Synscapes~\cite{wrenninge2018synscapes}, SmartPortraits~\cite{kornilova2022smartportraits}, and HOI4D~\cite{liu2022hoi4d} are treated as single views. To train on single-view datasets with a specified context length, we construct sequences by stacking independent views to the desired context length, and importantly always reset the state to $\bm{s}_0$ after each view. This allows us to jointly train using both multi-view and single-view data within the same batch. Although both EDEN~\cite{le2021eden} and SmartPortraits~\cite{kornilova2022smartportraits} provide camera poses,  EDEN~\cite{le2021eden} lacks clear documentation of camera conventions, and SmartPortraits~\cite{kornilova2022smartportraits} offers camera poses that are not synchronized with RGBD frames. Therefore, we treat both as single-view datasets.

For PointOdyssey~\cite{zheng2023pointodyssey}, we filter scenes with incorrect depth annotations~(mostly scenes with fogs, like \texttt{cab\_h\_bench\_ego2})
 and scenes with unrealistic motion and material~(like \texttt{Ani}). For BEDLAM~\cite{black2023bedlam}, we remove scenes with panorama backgrounds. 

\section{More Implementation Details}

\paragraph{Sequence Sampling Details.}
Our training dataset comprises a combination of video sequences and unordered photo collections. For video sequences, we subsample frames at intervals randomly selected between $1$ and $k$, where $k$ is set for each dataset based on its frame rate and camera motion. Within each sequence, either variable or fixed intervals are used, each accounting for approximately half of the samples.
For photo collections, we use similar methods as in DUSt3R~\cite{wang2024dust3r} and compute the overlap ratios between images to guide the frame sampling. 
Additionally, when the scene from a video is largely static, we shuffle the frames and treat them as a photo collection to increase data diversity. When the sequences contain major dynamic objects~(like sequences from  BEDLAM~\cite{black2023bedlam} and PointOdyssey~\cite{zheng2023pointodyssey} datasets), we only treat them as videos and feed frames into the model in temporal order using a fixed interval.

When the data is metric scale, frames (excluding the first frame) in a sequence are randomly masked with a 20\% probability and replaced by their corresponding raymap inputs, using ground truth intrinsics and poses. Note that raymap mode is activated only when data are in metric scale, as our model learns metric-scale 3D scene priors. When the 3D annotation is at an unknown scale, raymap querying is disabled to avoid scale inconsistency with the scene content captured in the state.

\begin{table*}[t]
\centering
\renewcommand{\arraystretch}{1.02}
\renewcommand{\tabcolsep}{1.5pt}
\resizebox{0.85\textwidth}{!}{
\begin{tabular}{@{}llcc>{\centering\arraybackslash}p{1.5cm}>{\centering\arraybackslash}p{1.5cm}|>{\centering\arraybackslash}p{1.5cm}>{\centering\arraybackslash}p{1.5cm}|>{\centering\arraybackslash}p{1.5cm}>{\centering\arraybackslash}p{1.5cm}|>{\centering\arraybackslash}p{1.2cm}@{}}
\toprule
 &  &  &  & \multicolumn{2}{c}{\textbf{Sintel}} & \multicolumn{2}{c}{\textbf{BONN}} & \multicolumn{2}{c}{\textbf{KITTI}} & \\ 
\cmidrule(lr){5-6} \cmidrule(lr){7-8} \cmidrule(lr){9-10}
\textbf{Alignment} & \textbf{Method} & \textbf{Optim.} & \textbf{Onl.\ } & {Abs Rel $\downarrow$} & {$\delta$\textless $1.25\uparrow$} & {Abs Rel $\downarrow$} & {$\delta$\textless $1.25\uparrow$} & {Abs Rel $\downarrow$} & {$\delta$ \textless $1.25\uparrow$} & \textbf{FPS} \\ 
\midrule
\multirow{13}{*}{\begin{minipage}{3cm}Per-sequence \\
scale  \& shift\end{minipage}} 
 & Marigold~\cite{marigold} &   &\checkmark & 0.532 & {51.5} & {0.091} & {93.1} & {0.149} & {79.6} & $<$0.1 \\
  & Depth-Anything-V2~\cite{depth_anything_v2} &  &  \checkmark& 0.367 & {55.4} & {0.106} & {92.1} &{0.140} & {80.4} & 3.13 \\
   & NVDS~\cite{NVDS} &  & \checkmark & 0.408 & {48.3} & {0.167} & {76.6} & {0.253} & {58.8} & - \\
    & ChronoDepth~\cite{shao2024learning} &  & \checkmark & 0.687 & {48.6} & {0.100} & {91.1} & {0.167} &{75.9} & 1.89 \\
     & DepthCrafter~\cite{hu2024-DepthCrafter} &  &  \checkmark& \textbf{0.292} & \textbf{{69.7}} & {0.075} & \textbf{{97.1}} & \underline{{0.110}} & \underline{{88.1}} & 0.97 \\
      & Robust-CVD~\cite{robustcvd} &  & \checkmark & 0.703 & {47.8} & {-} & {-} & - & - & - \\
      & CasualSAM~\cite{casualsam} & \checkmark  & & 0.387 & {54.7} & {0.169} & {73.7} & {0.246} & 62.2 & - \\
& DUSt3R-GA~\cite{wang2024dust3r} &  \checkmark & & 0.531 & {51.2} & {0.156} & {83.1} & {0.135} & {81.8} & 0.76 \\
& MASt3R-GA~\cite{leroy2024mast3r} &   \checkmark & & \underline{0.327} & \underline{59.4} & {0.167} & {78.5} & {0.137} & {83.6} & 0.31 \\

& MonST3R-GA~\cite{zhang2024monst3r} &  \checkmark &  & {0.333} & {59.0} & \textbf{{0.066}} & \underline{96.4} & {0.157} & {73.8} & 0.35 \\
& Spann3R~\cite{spann3r} &  & \checkmark & 0.508 & {50.8} & {0.157} & {82.1} & {0.207} & {73.0} &13.55 \\

& \ourmethod &  & \checkmark & 0.454  & 55.7 & \underline{0.074} & {94.5} & \bf{0.106} & \bf{88.7} &  16.58 \\

\midrule
\multirow{6}{*}{\begin{minipage}{3cm}Per-sequence scale\end{minipage}} & DUSt3R-GA~\cite{wang2024dust3r} &  \checkmark & & 0.656 & {45.2} & {0.155} & {83.3} & \underline{0.144} & \underline{81.3} & 0.76 \\
& MASt3R-GA~\cite{leroy2024mast3r} &   \checkmark & & 0.641 & {43.9} & {0.252} & {70.1} & {0.183} & {74.5} & 0.31 \\

& MonST3R-GA~\cite{zhang2024monst3r} &  \checkmark &  & \textbf{0.378} & \textbf{55.8} & \textbf{0.067} & \textbf{96.3} & {0.168} & {74.4} & 0.35 \\
& Spann3R~\cite{spann3r} &  & \checkmark & 0.622 & {42.6} & {0.144} & {81.3} & {0.198} & {73.7} &13.55 \\

& \ourmethod &  & \checkmark & \underline{0.421}  & \underline{47.9} & \underline{0.078} & \underline{93.7} & \textbf{0.118} & \textbf{88.1} &  16.58 \\

\midrule
 \multirow{2}{*}{\begin{minipage}{3cm}Metric scale\end{minipage}} & MASt3R-GA~\cite{leroy2024mast3r} & \checkmark &  & \bf 1.022  & 14.3 & 0.272 & 70.6 & 0.467 & 15.2  & 0.31 \\  
& \ourmethod &  & \checkmark & {1.029} & \textbf{23.8} & \textbf{0.103} & \textbf{88.5}& \textbf{0.122} & \textbf{85.5}  &  16.58 \\
\bottomrule
\end{tabular}
}
\caption{\small{
\textbf{Video Depth Evaluation}. We report scale\&shift-invariant depth, scale-invariant depth and metric depth accuracy on Sintel, Bonn, and KITTI datasets. Methods requiring global alignment are marked ``GA'', while ``Optim.'' and ``Onl.'' indicate optimization-based and online methods, respectively. We also report the FPS on KITTI dataset using 512$\times$ 144 image resolution for all methods, except \spanner{} which only supports 224$\times$224 inputs.
}
}
\label{tab:video_depth_supp}
\end{table*}

\begin{table*}[t]
\centering
\footnotesize
\renewcommand{\arraystretch}{1.}
\renewcommand{\tabcolsep}{2.5pt}
\resizebox{0.9\textwidth}{!}{
\begin{tabular}{@{}clccccc|ccc|ccc@{}}
\toprule
& & & & \multicolumn{3}{c}{\textbf{Sintel}} & \multicolumn{3}{c}{\textbf{TUM-dynamics}} & \multicolumn{3}{c}{\textbf{ScanNet}} \\ 
\cmidrule(lr){5-7} \cmidrule(lr){8-10} \cmidrule(lr){11-13}
{} & {\textbf{Method}}  &  \textbf{Optim.} & \textbf{Onl.} & {ATE $\downarrow$} & {RPE trans $\downarrow$} & {RPE rot $\downarrow$} & {ATE $\downarrow$} & {RPE trans $\downarrow$} & {RPE rot $\downarrow$} & {ATE $\downarrow$} & {RPE trans $\downarrow$} & {RPE rot $\downarrow$} \\ 
\midrule
& DROID-SLAM~\cite{teed2021droid} & &\checkmark & 0.175 & 0.084 & \underline{1.912} & - & - & - & - & - & - \\ 
& DPVO~\cite{teed2023deep} & & \checkmark & \underline{0.115} & \underline{0.072} & 1.975 & - & - & - & - & - & - \\ 
& LEAP-VO~\cite{chen2024leap} & & \checkmark & \textbf{{0.089}} & \textbf{0.066} & \textbf{1.250} & {{0.068}} & {0.008} & {1.686} & {{0.070}} & {{0.018}} & {{0.535}} \\ 
\midrule
& Particle-SfM~\cite{zhao2022particlesfm} & \checkmark & & \underline{0.129} & {0.031} & \bf{0.535} & - & - & - & 0.136 & 0.023 & 0.836 \\ 

& Robust-CVD~\cite{robustcvd} & \checkmark & & 0.360 & 0.154 & 3.443 & 0.153 & 0.026 & 3.528 & 0.227 & 0.064 & 7.374 \\ 
 & CasualSAM~\cite{casualsam} & \checkmark & & 0.141 & \textbf{0.035} & \underline{0.615} & \underline{0.071} & \textbf{0.010} & 1.712 & 0.158 & 0.034 & 1.618 \\ 
 & DUSt3R-GA~\cite{wang2024dust3r} & \checkmark & & 0.417 & 0.250 & 5.796 & 0.083 & 0.017 & 3.567 & {0.081} & 0.028 & 0.784 \\  
{ }& MASt3R-GA~\cite{leroy2024mast3r} & \checkmark & & {{0.185}} & {0.060} & {1.496} & {\bf{0.038}} & {\underline{0.012}} & {\bf{0.448}} & {\underline{0.078}} & {\underline{0.020}} & {\bf {0.475}} \\ 
& MonST3R-GA~\cite{zhang2024monst3r}  & \checkmark & & \bf {{0.111}} & \underline{0.044} & {0.869} & {{0.098}} & {{0.019}} & {\underline{0.935}} & {\bf{0.077}} & {\bf{0.018}} & {\underline{0.529}} \\ 
\midrule
&  \duster{}~\cite{wang2024dust3r} & &\checkmark  & {\underline{0.290}} & {0.132} & {7.869} & {{0.140}} & {{0.106}} & {{3.286}} & {{0.246}} & {{0.108}} & {{8.210}} \\ 

& Spann3R~\cite{spann3r} & &\checkmark  & {{0.329}} & \underline{0.110} & \underline{4.471} & {\underline{0.056}} & {\underline{0.021}} & {\underline{0.591}} & {\bf{0.096}} & {\underline{0.023}} & {\underline{0.661}} \\ 

& Ours &  & \checkmark & \bf 0.213 & \bf 0.066 & \bf 0.621 & \textbf{0.046} & \bf 0.015 & \bf 0.473 & \underline{0.099} & \textbf{0.022} & \textbf{0.600}\\  
\bottomrule
\end{tabular}
}
\caption{\textbf{Evaluation on Camera Pose Estimation} on Sintel~\cite{sintel}, TUM-dynamic~\cite{tum-dynamics}, and ScanNet~\cite{dai2017scannet} datasets.
Note that unlike the the rest of the methods, the three methods in the first section require ground truth camera intrinsics as input.
}
\label{tab:video_pose_supp}
\end{table*}

\vspace{0.5em}
\noindent\textbf{More Architecture Details.} 
Similar to DUSt3R~\cite{wang2024dust3r}, we reduce training costs by first training the model on $224\times224$ image resolution with linear heads, and then increasing the resolution and setting the longer side of the images to 512 pixels. Specifically, in the first two stages of training, $\headself$ and $\headcross$ are implemented as linear layers. In the final two stages, $\headself$ and $\headcross$ are switched to DPT~\cite{Ranftl2021} architecture. 
Compared to $\headself$, $\headcross$ incorporates an additional modulation function, which modulates $\fts$ using the pose token $\zts$ within the Layer Normalization layers. This modulation design is inspired by LRM~\cite{hong2023lrm} and aims to integrate pose information to achieve implicit rigid transformations. Specifically, within $\headcross$, we first use two self-attention blocks modulated by the pose token $\zts$ to generate the pose-modulated tokens, which is then fed as input to either the linear or DPT architecture to generate the final pointmap output $\pmcross$. The dimension of $\zts$ is $768$, and $\headpose$ is a 2-layer MLP whose hidden size is $768$. We apply Rotary Positional Embedding~(ROPE)~\cite{su2024roformer} to the query and key feature before each attention operation.

\paragraph{More Training Details.}
In the first stage of training, we use the following datasets: ARKit, ARKit-HighRes, ScanNet, ScanNet++, TartanAir, Waymo, MapFree, BlendedMVS, HyperSim, MegaDepth, Unreal4K, DL3DV, CO3Dv2, WildRGBD, and VirtualKITTI2. In the second stage, we incorporate the rest of datasets. In the final stage (long context training), we exclude single-view datasets (EDEN, IRS, Synscapes, 3D Ken Burns, SmartPortraits, UrbanSyn, and HOI4D) and train only on multi-view datasets, as the goal of the final stage training is to enhance scene-level reasoning within a sequence. Unlike \duster{}, which applies color jittering to each image independently, we perform sequence-level color jittering by applying the same color jitter across all frames in a sequence.

\section{More Comparisons}

\paragraph{Video Depth Estimation.}
We expand the video depth comparison in the main paper and compare with a wider range of baseline methods, including single-frame depth techniques (Marigold \cite{marigold} and Depth-Anything-V2 \cite{depth_anything_v2}), video depth approaches (NVDS \cite{NVDS}, ChronoDepth \cite{shao2024learning}, and DepthCrafter \cite{hu2024-DepthCrafter}), and joint depth-and-pose methods such as Robust-CVD \cite{robustcvd}, CasualSAM \cite{casualsam}, DUSt3R \cite{wang2024dust3r}, MASt3R \cite{leroy2024mast3r}, MonST3R \cite{zhang2024monst3r}, and Spann3R \cite{spann3r}. The results are shown in Tab.~\ref{tab:video_depth_supp}.

\vspace{0.5em}
\noindent\textbf{Camera Pose Estimation}
Similar to video depth estimation, we include a diverse set of baselines for camera pose estimation. Learning-based visual odometry methods, such as DROID-SLAM \cite{teed2021droid}, DPVO \cite{teed2023deep}, and LEAP-VO \cite{chen2024leap}, require ground truth camera intrinsics as input. Optimization-based methods, including Particle-SfM \cite{zhao2022particlesfm}, Robust-CVD \cite{robustcvd}, CasualSAM \cite{casualsam}, DUSt3R-GA \cite{wang2024dust3r}, MASt3R-GA \cite{leroy2024mast3r}, and MonST3R-GA \cite{zhang2024monst3r}, generally operate more slowly compared to online methods like Spann3R \cite{spann3r} and our proposed approach. To assess performance in an online setting, we also evaluate DUSt3R without global alignment. The results are presented in Tab.~\ref{tab:video_pose_supp}.

\end{document}